\documentclass[11pt,a4paper]{article}
\usepackage[hyperref]{acl2019}
\usepackage{times}
\usepackage{latexsym}

\usepackage{url}

\usepackage[colorinlistoftodos]{todonotes}  
\usepackage{algorithmic}
\usepackage{stmaryrd} 
\usepackage[algo2e,linesnumbered,ruled]{algorithm2e} 
\usepackage{multirow}
\usepackage{amsmath}
\usepackage{comment} 
\usepackage{epstopdf} 

\makeatletter
\newcommand{\removelatexerror}{\let\@latex@error\@gobble}
\makeatother

\newcommand{\ignore}[1]{} 

\aclfinalcopy 

\title{Zero-Shot Semantic Parsing for Instructions}

\author{Ofer Givoli\textsuperscript{1} \and Roi Reichart\textsuperscript{2}\\
  \textsuperscript{1} Faculty of Computer Science, Technion, IIT \\
  \textsuperscript{2} Faculty of Industrial Engineering and Management, Technion, IIT \\
  {\tt ogivoli@cs.technion.ac.il, roiri@technion.ac.il} \\}

\date{}

\SetKwProg{Fn}{Function}{}{} 

\begin{document}
\maketitle
\begin{abstract}

We consider a zero-shot semantic parsing task: parsing instructions into compositional logical forms, in domains that were not seen during training. We present a new dataset with 1,390 examples from 7 application domains (e.g. a calendar or a file manager), each example consisting of a triplet: (a) the application's initial state, (b) an instruction, to be carried out in the context of that state, and (c) the state of the application after carrying out the instruction. We introduce a new training algorithm that aims to train a semantic parser on examples from a set of source domains, so that it can effectively parse instructions from an unknown target domain. We integrate our algorithm into the floating parser of \newcite{pasupat2015compositional}, and further augment the parser with features and a logical form candidate filtering logic, to support zero-shot adaptation. Our experiments with various zero-shot adaptation setups demonstrate substantial performance gains over a non-adapted parser.\footnote{Our code and data are available at: \href{https://github.com/givoli/TechnionNLI}{https://github.com/givoli/TechnionNLI}.}
\end{abstract}

\section{Introduction}
\label{ChapOrSec:Introduction}
The idea of interacting with machines via natural language instructions and queries has fascinated researchers for decades \cite{winograd1971procedures}. Recent years have seen an increasing number of applications that have a natural language interface, either in the form of chatbots or via ``intelligent personal assistants'' such as Alexa (Amazon), Google Assistant, Siri (Apple), and Cortana (Microsoft).

In the near future, we may find ourselves in a world where even more functionality could be accessed via a natural language user interface (NLUI). If so, we better seek answers to the following questions: 
Will every developing team need to hire NLP experts to develop a NLUI for their specific application? Can we hope for a general framework that once trained on annotated data from a set of domains, does not require annotated data from a newly presented domain? Previous work on tasks related to NLUI for applications mostly relied on in-domain data (e.g. ~\newcite{artzi2013weakly,long2016simpler}), and papers that did not rely on in-domain data did not attempt to parse instructions into compositional logical forms \cite{kimnatural__year_2016}.

To fill this gap, we address the task of \textit{zero-shot semantic parsing for instructions}: training a parser so that it can parse instructions into compositional logical forms, where the instructions are from domains that were not seen during training. Formally, our task assumes a set $D=\{d_1, ..., d_n\}$  of source domains, each corresponding to a simple application (e.g. a calendar or a file manager) and an application program interface (API) consisting of a set of \textit{interface methods}. Each interface method is augmented with a list of \textit{description phrases} that are expected to be used by the users of the application to ask for the invocation of that method. These instructions are to be parsed into logical forms that denote a method call with specific arguments.

\begin{figure}
	\centering
	\includegraphics[width=0.45\textwidth]{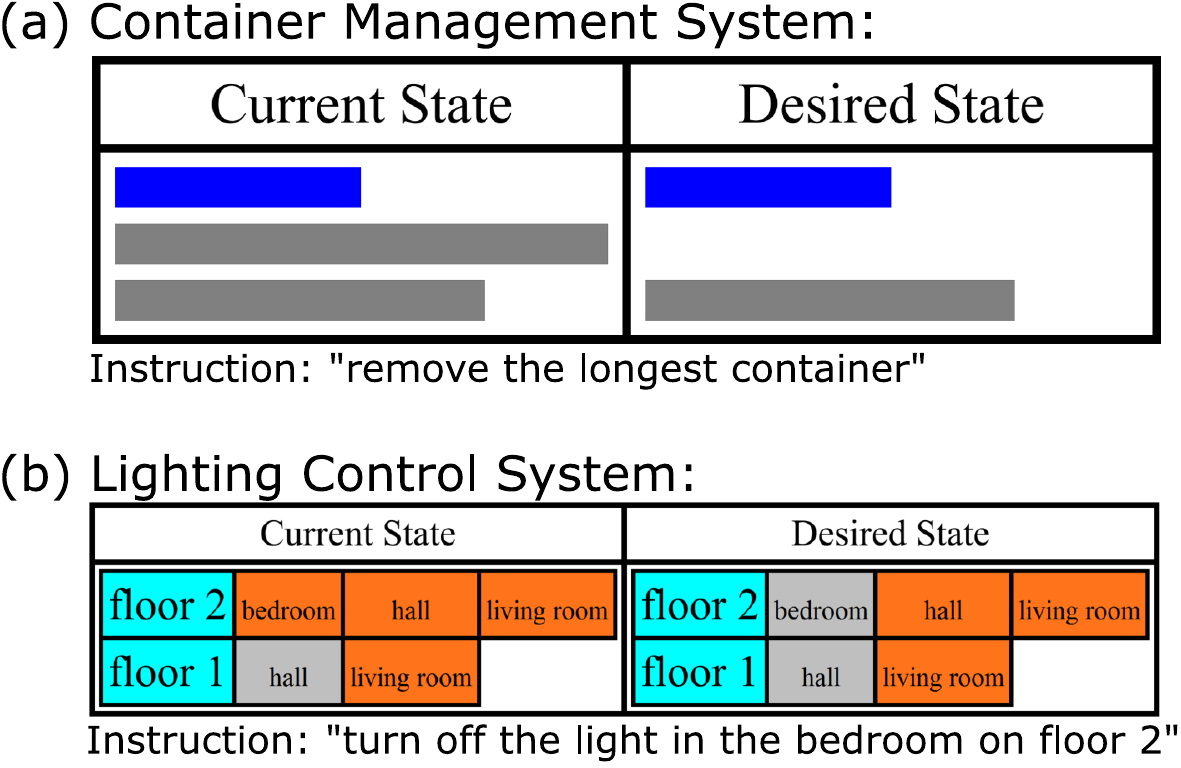}
	\caption{Two examples from our dataset.  Annotators are presented with a visualization of initial and desired states, and are asked to write an instruction that will transfer the system from the former state to the latter. \textbf{(a)} Shipping containers can be either empty (gray) or full (blue), and have varying length. \textbf{(b)} Rooms can have their lights on (orange) or off (gray).}
	
    \label{fig:IntroHIT}
\end{figure}

We collected a new dataset of 1,390 examples from 7 domains. Each example in the dataset is a triplet consisting of (a) the application's initial state, (b) an instruction, to be carried out in context of that state, and (c) the state of the application after carrying out the instruction, also referred to as the \emph{desired~state}. The instructions were provided by MTurk workers, one for each pair of initial and desired states. Figure \ref{fig:IntroHIT} demonstrates examples from two of the domains in~our~dataset. 

We present a new training algorithm for zero-shot semantic parsing, which involves learning the weights in two steps, such that in each step different source domains are used. Our training algorithm  is motivated by the goal of optimizing the weights for unseen domains rather than for the source domains, and is integrated into the floating parser  \cite{pasupat2015compositional}: a parser that was designed for question answering, but is easy to adjust to instruction execution (see \S~\ref{sec:prediction}). 

To further assist the parser in dealing with the zero-shot setup, we extract additional features, mostly based on co-occurrence of primitive logical forms and the description phrases that are provided for the interface methods. We also use the application logic to dismiss candidate logical forms that represent a method call which does not modify the application state or results in an exception being thrown from the application logic.

Our training algorithm yields an averaged accuracy of 44.5\%, compared to 39.1\% of the parser when trained with its original AdaGrad training algorithm \cite{duchi2011adaptive}, but with our features and filtering logic. Further exclusion of our features and filtering logic decreases accuracy to 28.3\%. We demonstrate that, relative to the baseline, our training algorithm yields smaller weights to some features in a way that can be expected to benefit previously unseen domains.

\section{Previous Work}
Previous work on executable semantic parsing can be classified as either work on question answering (e.g. ~\newcite{clarke2010driving,pasupat2015compositional,krishnamurthy2017neural}) or instruction parsing (e.g. ~\newcite{artzi2013weakly,long2016simpler}). The result of executing a logical form is either an answer or a change in some state, respectively. Our work is the first to address the novel semantic parsing task of mapping natural language instructions into compositional logical forms in zero-shot settings.

While in our task each example contains a single sentence instruction, there are works on semantic parsing for instruction sequences ~\cite{macmahon2006walk, long2016simpler}, but not in a zero-shot setup. We keep zero-shot parsing of instruction sequences for future research.

A lot of work has been done on slot tagging and goal-oriented dialog \cite{kimnatural__year_2016,gasic2016dialogue, zhao2018zero}  
which, similarly to our work, involves automatically enabling an NLUI to a given system. Unlike in our task, the tasks that are investigated in those papers do not require the mapping of natural language to a meaning representation over a space of compositional logical forms. Other tasks related to ours include program synthesis \cite{raza2015compositional,desai2016program} and mapping natural language to bash code \cite{lin2018nl2bash}, but these also did not consider zero-shot setups and did not synthesize code in the context of an application state.


\paragraph{Semantic Parsing with In-domain Data}

Among the semantic parsing work that relied on in-domain data, many relied on a domain-specific lexicon \cite{kwiatkowski2010inducing,gerber2011bootstrapping,krishnamurthy2012weakly,zettlemoyer2012learning,cai2013large} which maps natural language phrases to primitive logical forms. Many of these works automatically constructed a domain-specific lexicon using some additional domain-specific resources that are associated with the entities and relations of the given domain. Such a resource can be either a very large corpus ~\cite{gerber2011bootstrapping,krishnamurthy2012weakly}, search results from the Web \mbox{\cite{cai2013large}} or pairs of a sentence and an associated logical form ~\cite{zettlemoyer2012learning,kwiatkowski2010inducing}. 
In our task none of the above resources is assumed to be available but instead we use the description phrases of the interface~methods.
%
%

\paragraph{Cross-domain and Zero-shot Semantic Parsing}

Previous semantic parsers use supervised training, either with \cite{zettlemoyer2012learning,kwiatkowski2010inducing} or without \cite{clarke2010driving,berant2013semantic} logical forms annotation, in addition to unsupervised training ~\cite{goldwasser2011confidence}. We take the supervised approach with no logical form annotations.

While most semantic parsing work trains on in-domain data, 
there are some exceptions. \newcite{cai2013large} and \newcite{kwiatkowski2013scaling} introduced semantic parsers for question answering that can parse utterances from Free917 \cite{cai2013large} such that no Freebase entities or relations appear in both training and test examples. 
We also note the relevance of the dataset presented in \newcite{pasupat2015compositional} which contains questions about Wikipedia tables, such that the context of each question is a single table. They evaluate a parser on questions about tables that have not been observed during training.
Their work does not fully constitute zero-shot semantic parsing due to table columns across the train/test split that share column headers (which correspond to primitive logical forms that represent relations). Our parser is based on the floating parser introduced in that paper, and the space of logical forms we use is very similar to theirs (see \S~\ref{sec:prediction}).

Recently, \newcite{herzig2017neural} and \newcite{su2017cross} experimented with the Overnight dataset \cite{wang2015building} in cross-domain settings. These papers did not experiment with zero-shot setups (i.e. training without any data from the target domain), and they both observed that the less in-domain training data was used, the more training data from other domains was valuable. Recently \newcite{herzig2018decoupling} explored zero-shot semantic parsing with the Overnight dataset. Their framework, unlike ours, requires logical form annotation, and is designed for question answering rather than instruction parsing. 

Another zero-shot semantic parsing task was introduced in \newcite{yu2018spider}. The task requires mapping natural language questions to SQL queries, and includes a setting in which no databases appear in both the training and test sets (as attempted in \newcite{yu2018syntaxsqlnet}).

\section{Task and Data}
We now describe our task and dataset.

\subsection{Task} \label{subsec-task}
Our task involves parsing a natural language instruction, in the context of a small application, into a method call that corresponds to the application's API. For example, the \textsc{Lighting} domain corresponds to a lighting control system application that allows the user to turn the lights on and off in each room in their house.

Formally, a domain has a set of \textit{interface methods} (e.g. $\mathrm{turnLightOn}$ and $\mathrm{turnLightOff}$) that can be invoked with some arguments. Each argument is a set of entities (e.g. a set of $\mathrm{Room}$ entities). There are two kinds of \textit{entity types}: \mbox{domain-specific} (e.g. $\mathrm{Room}$) and non \mbox{domain-specific} ($\mathrm{Integer}$, $\mathrm{String}$). Each interface method is augmented with 1-3 description phrases that correspond to its functionality. For example, the interface method $\mathrm{removeEvents}$ from the \textsc{Calendar} domain has the description phrases \textit{remove} and \textit{cancel}.

A \textit{state} defines a knowledge base, consisting of a set of $(e_1,r,e_2)$ triplets, where $e_1$ and $e_2$ are entities and $r$ is a relation. For example, a knowledge base in the \textsc{Lighting} domain might contain the triplet $(\mathrm{room3}, \mathrm{floor}, 2)$ which indicates that $\mathrm{room3}$ is on the second floor of the house. In figure \ref{fig:IntroHIT} (b) we demonstrate two possible states in the \textsc{Lighting} domain. In the first one, there is a bedroom on the second floor with the lights turned on. If that room is represented in the state $s$ by the entity $\mathrm{room1}$, the following triplets will be in the knowledge base of $s$: $(\mathrm{room1},\mathrm{name},\mathrm{bedroom})$, $(\mathrm{room1},\mathrm{floor},2)$ and $(\mathrm{room1},\mathrm{lightMode},\mathrm{ON})$.

Our task is limited to mapping an utterance into a single \textit{method call}. A method call formally consists of an [interface method, argument list] pair. The invocation of the method call changes the application state according to the \mbox{deterministic application logic}.

Our dataset consists of examples from 7 application domains. Each example is a triplet $(s,x,s')$, where $s$ is an initial application state, $x$ is a natural language instruction and $s'$ is a desired application state, resulting from carrying out the instruction $x$ on the state $s$. The task is to train a parser with examples from a given subset of domains (the source domains), so that it can effectively parse instructions from a different domain (the target domain), which is unseen at training.

\subsection{Data} \label{subsec:data}

Our task requires a dataset that consists of examples from multiple domains, such that each example corresponds to an instruction in the context of an application state. 
Following \newcite{long2016simpler}, we constructed the dataset by presenting human annotators with visualizations of initial and desired state pairs. The annotators were then asked to write an English instruction that can be executed in order to transfer the application from the initial state to the desired state (see figure \ref{fig:IntroHIT}). 

Given a domain and an interface method, we randomly generate a state pair with the \mbox{following steps:}
\begin{enumerate}
\item Randomly generating an initial state. For example, in the \textsc{Lighting} domain (see figure
1 (b) of the main paper), we randomly select the number of floors, number of rooms in each floor, and for each room we randomly select a name (e.g. $bedroom$) from a list of possible names, and a light mode (either \mbox{$\mathrm{ON}$ or $\mathrm{OFF}$).}
\item Randomly selecting arguments for the interface methods. For example, in the \textsc{Lighting} domain we randomly select a set of rooms as an argument for the interface method ($\mathrm{turnLightOn}$ or $\mathrm{turnLightOff}$).
\item Invoking the interface method with the selected arguments on the initial state. If the result is a state that is identical to the initial state, or if an error occurred during execution, we go back to step 2. After 1,000 failed attempts we deem the random initial state as problematic and go back to step 1.
\end{enumerate}
With this process, we collected 1,390 examples from 7 domains (Table \ref{tab:DomainTable}). We used the MTurk platform and recruited annotators located in the US with at least 1,000 approved tasks and a task approval rate of at least 95\%. Our dataset contains utterances written by 53 unique annotators.

Throughout the dataset construction we blocked 16 annotators that generated utterances that did not correspond to our instructions (mostly, referring to irrelevant details of the provided figure which are not part of the domain represented by the visualized states). Annotators were paid 15-23 cents per task (i.e. per utterance they write given a state~pair). 

Each initial and desired state pair was given to a single annotator. We filtered out examples with utterances that consist of more than one sentence (we instructed annotators to write only one sentence). The average instruction length in the training set is 8.1 words. 

\begin{table*}[t!]
\begin{center}
\resizebox{\textwidth}{!}{
\begin{tabular}{|c|p{8cm}|p{8.5cm}|c|c|c|}
\hline
Domain & Domain-specific entity types & Interface method (parameters in parentheses) & Examples \# & Train \# & Test \# \\
\hline\hline
	\textsc{Calendar} &  \textbf{Event} $($title, startTime, location, color, attendees$)$ &  \textbf{removeEvents}(Collection\textless Event\textgreater)\newline \textbf{setEventColor}(Collection\textless Event\textgreater, Color)  & 199 & 100 & 99  \\
    \hline
    $\begin{matrix} \text{\textsc{Container}} \\ \text{(container management system)} \end{matrix}$ &  \textbf{ShippingContainer} $($length,\newline contentState $\in\{\text{LOADED},\text{UNLOADED}\})$  &  \textbf{loadContainers}(Collection\textless ShippingContainer\textgreater)\newline \textbf{unloadContainers}(Collection\textless ShippingContainer\textgreater)\newline \textbf{removeContainers}(Collection\textless ShippingContainer\textgreater)  & 201 & 104 & 97  \\
    \hline
    $\begin{matrix} \text{\textsc{File}} \\ \text{(file manager)} \end{matrix}$ &  \textbf{Directory} $($name, childFiles, childDirectories$)$\newline \textbf{File} $($name, type, sizeInBytes$)$&  \textbf{removeFiles}(Collection\textless File\textgreater)\newline \textbf{moveFiles}(Collection\textless File\textgreater,Directory) & 194 & 100 & 94  \\
    \hline
    $\begin{matrix} \text{\textsc{Lighting}} \\ \text{(lighting control system)} \end{matrix}$ &  \textbf{Room} $($name, floor,\newline lightMode $\in\{\text{ON},\text{OFF}\})$&  \textbf{turnLightOn}(Collection\textless Room\textgreater)\newline \textbf{turnLightOff}(Collection\textless Room\textgreater) & 209 & 104 & 105  \\
    \hline
	\textsc{List} &  \textbf{Element} $($value$)$ &  \textbf{remove}(Collection\textless Element\textgreater)\newline \textbf{moveToBeginning}(Element)\newline \textbf{moveToEnd}(Element) & 202 & 102 & 100  \\
	\hline
	\textsc{Messenger} &  \textbf{User} $($firstName$)$\newline \textbf{ChatGroup} $($contacts, muted, participantsNumber$)$ &  \textbf{createChatGroup}(Collection\textless User\textgreater)\newline \textbf{deleteChatGroups}(Collection\textless ChatGroup\textgreater)\newline \textbf{muteChatGroups}(Collection\textless ChatGroup\textgreater)\newline \textbf{unmuteChatGroups}(Collection\textless ChatGroup\textgreater) & 186 & 96 & 90  \\
	\hline
	$\begin{matrix} \text{\textsc{Workforce}} \\ \text{(workforce management system)} \end{matrix}$ &  \textbf{Employee} $($name, manager, salary,\newline position $\in\{\text{DEVELOPER},\text{QA}, \text{MANAGER}\})$ &  \textbf{assignEmployeesToNewManager}(\newline Collection\textless Employee\textgreater, Employee)\newline \textbf{fireEmployees}(Collection\textless Employee\textgreater)\newline \textbf{assignEmployeeToNewPosition}(Employee, Position)\newline \textbf{updateSalary}(Employee, int)\newline & 199 & 101 & 98  \\
	\hline
\end{tabular}
}
\end{center}
\caption[The domains in our dataset.]{The domains in our dataset. The interface method parameters are presented in Java syntax. In the second column, the properties of the non-primitive entities appear in parentheses. 
}
\label{tab:DomainTable}
\end{table*}

We consider 7 domains: (a) \textsc{Calendar}: removing calendar events and setting their color;
(b) \textsc{Container}: loading, unloading and removing shipping containers; (c) \textsc{File}: removing files and moving them from one directory to another; (d) \textsc{Lighting}: turning lights on and off in rooms inside a house; (e) \textsc{List}:  removing elements and moving an element to the beginning/end of a list;
(f) \textsc{Messenger}: creating/deleting chat groups and muting/unmuting them; and (g) \textsc{Workforce}: assigning employees to a new manager, firing employees, assigning an employee to a new position and updating an employee's salary.


Our choice of domains aims to include a variety of  linguistic phenomena. These include superlatives (e.g. \textit{remove the longest container} in \textsc{Container}, figure \ref{fig:IntroHIT} (a)), spatial language (e.g. \textit{turn off the light in the bedroom on floor 2} in \mbox{\textsc{Lighting}}, figure \ref{fig:IntroHIT} (b)), and temporal language (e.g. \textit{delete my last two appointments on Thursday}, from \textsc{Calendar}). Also, the domains are chosen to be rich enough to allow utterances with highly compositional logical forms. 


\section{Zero-Shot Semantic Parsing For Instructions} \label{ChapOrSec:Approach}

We modify the floating parser (henceforth denoted with \textit{FParser}), to address zero-shot learning in three ways: (a) presenting a new training algorithm; (b) filtering logical form candidates based on the application logic; and (c) adding new features. We begin with a brief description of the FParser and then go on to describe our approach.

\subsection{The Floating Parser} \label{sec:prediction}

The FParser was designed to handle unseen predicates, in the context of answering questions about Wikipedia tables that did not appear during training.\footnote{Their task diverged from zero-shot settings due to column headers that appear both in training and test tables.} It is hence a natural starting point for our zero-shot setup. We now describe the FParser and its inference algorithm (with necessary model modifications to support instruction parsing).

For each inference, the input of the parser is an initial application state $s$, a set of interface methods and their description phrases, and a natural language instruction $x$. The state $s$ defines a knowledge base $K_s$ of $(entity,relation,entity)$ triples. The parser generates a set of logical form candidates $Z_x$ that can be executed over the knowledge base $K_s$ to produce a \textit{method call} $c$ formulated as an (interface method, argument list) pair. The method call $c$ can be invoked in the context of $s$ with the provided application logic, producing the \textit{denotation} $y = \llbracket z\rrbracket_s$, the resulting state.

For each logical form $z\in Z_x$ the parser extracts a feature vector $\phi(x,s,z)$. The probability assigned to a logical form candidate $z \in Z_x$ is defined by a log-linear model:
$$p_\theta(z|x,s)=\frac{\exp(\theta^T\phi(x,s,z))}{\sum_{z'\in Z_x}{\exp(\theta^T\phi(x,s,z'))}}\label{eq:modeled_probability}$$
where $\theta$ is the weight vector. The logical form with maximal probability is chosen as the predicted logical form, and its denotation is the predicted desired state. Our logical form space is based on $\lambda$-DCS \cite{liang2013lambda}, as in the original FParser, but we use an additional derivation rule that derives the logical form $f(z_1,...,z_n)$, denoting a method call, given the primitive logical form $f$ (denoting an interface method) and the logical forms $z_1,...,z_n$ (each denoting a set of entities that correspond to an argument of $f$).


The objective function is the L1 regularized log-likelihood of the correct denotations across the training data:
\begin{equation}
\label{eq:objective}
J(\theta)=\frac{1}{N}\sum_{i=1}^N{\log p_\theta(y_i|x_i,s_i)-\lambda\|\theta\|_1}
\end{equation}
where $p_\theta(y|x,s)$ is the sum of the probabilities assigned to all the candidate logical forms with the~denotation~$y$.


\subsection{Zero-Shot Parsing}
\label{sec:zero}

We now present our modified FParser. We start with our training algorithm, and proceed to the application logic filtering and our new features.

\subsubsection{Training: Gap Minimization via Domain Partitioning (GMDP)} \label{subsec:Training}


We start with some notations and definitions.
Let us denote the set of training domains with $D = d_1,...,d_{n}$. Let ${D=D_1\cup D_2}$ be some partition of the set $D$. The target domain is denoted with $d_{n+1}$. Let us now describe the training algorithm, formulated in figure \ref{fig:2step_algorithm}.

The \textsc{GMDP} algorithm consists of two steps. In the first step, an initial estimate of the model parameters, denoted with $\theta_{D_1}$, is learned on the training examples from the source domain subset $D_1$. $\theta_{D_1}$ is then used as an initialization for the second step, in which the parser is re-trained, this time on the training examples of the domains in $D_2$. 

In each of the two steps we update the weights with AdaGrad \cite{duchi2011adaptive}, the training algorithm of the original FParser, using the objective function in~equation~\ref{eq:objective}. Since this objective is non-convex and hence sensitive to its starting point (parameter initialization), the weights learned at the first step ($\theta_{D_1}$) have an impact on the final parameters of the parser ($\theta_{D_1,D_2}$).

The motivation of \textsc{GMDP} training is simple. A good zero-shot parser should perform well on examples from domains that have not been available to it during training. To address this challenge, this two step method first estimates its parameters with respect to one set of domains ($D_1$) and then adjusts those parameters to fit a second set of domains ($D_2$) that have not been available at the first training step. We refer to this adjustment process as \textit{gap minimization}.

The parser parameters learned by \textsc{GMDP} strongly depend on the domains included in $D_1$ and $D_2$, and on the extent to which the adaptation from $D_1$ to $D_2$  mimics the adaptation from $D = D_1 \cup D_2$ to the target domain $d_{n+1}$. In this paper we treat the division of domains to the $D_1$ and $D_2$ subsets as a hyper-parameter and tune it together with the other hyper-parameters of the parser. Because this tuning process has to do with the important division of the training domains to $D_1$ and $D_2$, we detail it here as part of the description of the algorithm.

For every target domain $d_{n+1}$ we iterate over the training domains $D = d_1,...,d_{n}$ in a leave one out manner, each time holding out one of the domains $d_i \in D$, training on the other domains ($D_{\neg i}$) with various hyper-parameter configurations and testing on the training data of $d_i$. We consider as a hyper-parameter the order of the domains in $D$, assigning the first $M$ domains to $D_1$ and the rest to $D_2$ ($d_i$ is excluded from the ordered list), where $M$ is another hyper-parameter. The hyper-parameter configuration that works best in those $n$ iterations (achieves the highest average accuracy on the held-out domains) is the one selected for the training of the final parser for $d_{n+1}$. When a parser for $d_{n+1}$ is then trained, we increase by one the size of either $D_1$ or $D_2$, whichever is larger, because this way the ratio between the size of $D_1$ and $D_2$ is kept as similar as possible to the ratio during the hyper-parameter tuning.




\begin{figure}
\removelatexerror
\begin{algorithm2e}[H]
\caption{\textit{Gap Minimization via Domain Partitioning} (\textsc{GMDP})}
\KwIn{Domains partitioning: $\{D_1,D_2\}$}
\KwOut{Weight vector: $\theta_{D_1,D_2}$}
\begin{algorithmic}
    \STATE Initialize $\theta_0$
    \STATE $\theta_{D_1} \leftarrow AdaGrad(D_1,\theta_0)$
    \STATE $\theta_{D_1,D_2} \leftarrow AdaGrad(D_2,\theta_{D_1})$ \\
    \Return  $\theta_{D_1,D_2}$
\end{algorithmic}
\end{algorithm2e}
\caption{The \textsc{GMDP} algorithm. 
}
\label{fig:2step_algorithm}
\end{figure}

\subsubsection{Logical Form Filtering}
\label{sec:logic}

The Fparser is a bottom-up beam-search parser, in which a dynamic programming table is  filled with derivations. Each cell in the table corresponds to a derivation size and a logical form category, where the size is  defined as the number of rules applied when generating the logical form.

We add an additional stage to the inference step of the parser. In this stage we filter logical form candidates based on the application logic, which is part of the domain definition. This filtering stage dismisses incorrect candidate logical forms when they represent a method call $c$ that either does not modify the application state or results in an exception being thrown. To do that, $c$ is invoked on the initial state $s$ and if the result is a state identical to $s$, or if an exception has been thrown by the application logic, we dismiss the candidate logical form. This added stage is especially important for zero-shot settings, in which the application logic of the target domain does not have any impact on the learned weights.

As an example to application logic based filtering, consider the \textsc{Lighting} domain in which the lights in some rooms can be turned on and off. A method call that turns off the lights in rooms where the lights are already off does not change the application state, and in such cases the corresponding logical form will be dismissed.
In the \textsc{Workforce} domain, attempting to assign employees to report to an employee who is not a manager results in an exception being thrown.

\subsubsection{Features}
\label{sec:features}

Given a state, an instruction and a logical form, 
we extract the relevant features of the FParser (phrase-predicate co-occurrence features\footnote{We use the term \textit{predicate} here to be consistent with the terminology used by \newcite{pasupat2015compositional}, but in this work it also includes primitive logical forms denoting interface methods.} and missing-predicate features) and add our own features. We extract features based on the description phrases: the phrases that are provided for each interface method as part of the domain definition. 
The description phrases are used to extract additional phrase-predicate co-occurrence features and missing-predicate features.
For example, consider the utterance \textit{Delete the largest file} from the \textsc{File} domain, with the logical form: 
\begin{align*}
& \mathrm{removeFiles}(\mathrm{argmax}(\mathbf{R}[type].\mathrm{File},  \\ & \mathbf{R}[sizeInBytes])) 
\end{align*}

The interface method $\mathrm{removeFiles}$ has the description phrase \textit{delete}, which matches the phrase \textit{Delete} in the utterance, resulting in the extraction of the corresponding co-occurrence features. Conversely, when the parser considers logical forms that contain the method $\mathrm{moveFiles}$ instead of $\mathrm{removeFiles}$, it will extract features indicating that a match between the unigram \textit{Delete} and a primitive logical form is possible but does not occur in the candidate logical form. We note the resembles of this technique to the way \newcite{tafjord2018quarel} handled properties that were unseen during training, using a list of provided words that are associated with the property.


We also extract features that correspond to the size of a candidate logical form (the number of derivation rules applied).
The extracted features indicate that the logical form size is larger than $n$, for any $n\geq 2$. These features captures a domain-independent preference for simplicity.


\section{Experimental Setup}
\label{ChapOrSec:Experiments}

\paragraph{Models and Baselines}

We compare eight combinations of parsers. Four models use our \textsc{GMDP} training algorithm: (a) \textsc{GMDP}: the full model (\S~\ref{sec:zero}); (b) \textsc{GMDP$-$A}: where we do not apply the application logic filtering (\S~\ref{sec:logic}); (c) \textsc{GMDP$-$F}: where we do not use our features (\S~\ref{sec:features}); 
and (d) \textsc{GMDP$-$FA}: where we omit both our new features and the application logic filtering. The other four models are identical, but they use the original \textsc{AdaGrad} training. Note that \textsc{AdaGrad$-$FA} corresponds to the original FParser (with minimal modifications to support instruction parsing).

Notice that in some real-world settings our application logic filtering, which requires invoking interface methods hundreds of times per inference, might be impractical (e.g. if executing the interface methods is computationally intensive). This motivates us to consider the results of the ablated model that does not use the application logic.

\paragraph{Experiments}

In each experiment we train the parsers on examples from 6 application domains and test them on the remaining domain. Our evaluation metric is accuracy: the fraction of the test examples where a correct denotation (desired state) is predicted. For examples where multiple logical forms achieve maximum score, we consider the fraction that yields the desired state.

While in our main results we report the accuracy of the parsers on the target domain's test set, for the error and qualitative analyses we report the accuracy on the target domain's training set. We do that in order to avoid multiple runs on the test sets; we do not use the target domain's training set for other purposes (e.g. hyper-parameter tuning).
The average number of training and test examples per domain is 101 and 97.6, respectively.

\paragraph{Hyper-parameter tuning}

We use a grid search and leave-one-out cross-validation over the source domains to tune the hyper-parameters. We tune the following hyper-parameters: the L1 regularization coefficient, the initial step-size, the number of training iterations (for the \textsc{GMDP} algorithm: the number of training iterations in the second step), and for the \textsc{GMDP} algorithm also: the number and identity of training domains used in the first ($D_1$) and second ($D_2$) steps, and the number of training iterations during the first step. We use a beam size of 200 and limit the number of rule applications per derivation to 15. We provide more details about the values of the hyper-parameters we consider in the appendix.

\section{Results}

\begin{table*}
\begin{center}
\resizebox{\textwidth}{!}{
\begin{tabular}{|c||c|c|c|c|c|c|c||c|}
\hline
$\begin{matrix} \text{Training Algorithm} \\ \text{and Model}\end{matrix}$  & \textsc{Calendar} & \textsc{Container} & \textsc{File} & \textsc{Lighting} & \textsc{List} & \textsc{Messenger} & \textsc{Workforce}  & Avg.\\
\hline\hline
\textsc{GMDP} & 32.8 & \textbf{38.1} & \textbf{28.7} & \textbf{48.6 *} & \textbf{58.8 *} & \textbf{61.7 *} & \textbf{42.9 *} & \textbf{44.5} \\
\hline
\textsc{GMDP$-$F} & 36.4 & 25.8 & 21.3 & 11.4 & 33.5 & 37.2 & 22.8 & 26.9 \\
\hline
\textsc{GMDP$-$A} & 36.4 & 24.7 & 21.8 & 26.7 & 37.8 & 53.6 & 40.8 & 34.5 \\
\hline
\textsc{GMDP$-$FA} & \textbf{37.4} & 21.4 & 21.3 & 8.6 & 29.0 & 37.6 & 23.3 & 25.5 \\
\hline
\textsc{AdaGrad} & 
    $\begin{matrix} \text{34.8} \\ \text{(39.9)} \end{matrix}$ & 
	$\begin{matrix} \text{\textbf{38.1}} \\ \text{(36.1)} \end{matrix}$ & 
	$\begin{matrix} \text{28.0} \\ \text{(39.4)} \end{matrix}$ & 
	$\begin{matrix} \text{36.2} \\ \text{(83.8)} \end{matrix}$ & 
	$\begin{matrix} \text{44.5} \\ \text{(88.0)} \end{matrix}$ & 
	$\begin{matrix} \text{53.3} \\ \text{(73.3)} \end{matrix}$ & 
	$\begin{matrix} \text{38.8} \\ \text{(57.7)} \end{matrix}$ & 
	$\begin{matrix} \text{39.1} \\ \text{(59.7)} \end{matrix}$ \\
\hline
\textsc{AdaGrad$-$F} & 35.4 & 28.9 & 20.6 & 23.8 & 38.7 & 50.2 & 29.3 & 32.4 \\
\hline
\textsc{AdaGrad$-$A} & 33.8 & 22.7 & 21.6 & 26.7 & 51.3 & 53.3 & 40.8 & 35.8 \\
\hline
\textsc{AdaGrad$-$FA} & 
$\begin{matrix} \text{29.8} \\ \text{(39.9)} \end{matrix}$ & 
$\begin{matrix} \text{21.6} \\ \text{(35.1)} \end{matrix}$ & 
$\begin{matrix} \text{19.5} \\ \text{(33.5)} \end{matrix}$ & 
$\begin{matrix} \text{10.5} \\ \text{(79.0)} \end{matrix}$ & 
$\begin{matrix} \text{39.6} \\ \text{(81.0)} \end{matrix}$ & 
$\begin{matrix} \text{49.6} \\ \text{(72.2)} \end{matrix}$ & 
$\begin{matrix} \text{27.4} \\ \text{(55.1)} \end{matrix}$ & 
$\begin{matrix} \text{28.3} \\ \text{(56.5)} \end{matrix}$ \\ 
\hline
\end{tabular}}
\end{center}
\caption{Test set accuracy. In parenthesis: result for the in-domain setup (training with the 96-104 training examples of the target domain). \textsc{GMDP} results marked with * represent a statistically significant difference from \textsc{AdaGrad} ($\alpha=0.05$, using the paired bootstrap test \cite{efron1994introduction}, following \newcite{dror2018hitchhiker}), we do not mark the cell in the Average column as statistically significant due to it being the average of accuracy values from seven different distributions.
}
\label{tab:MainResults}
\end{table*}



Our results are summarized in Table \ref{tab:MainResults}. 
\textsc{GMDP} outperforms \textsc{AdaGrad$-$FA}, the original FParser, in all the domains, and by 16.2\% on average accuracy.
We next analyze the importance of each of our zero-shot components: the training algorithm, new features and \mbox{application logic filtering}.


\paragraph{Training Algorithms}

\textsc{GMDP} (our full model) yields an averaged accuracy of 44.5\%, outperforming \textsc{AdaGrad}, which is identical to our full model except that training is performed with AdaGrad,  by 5.4\%. In four domains \textsc{GMDP} outperforms \textsc{AdaGrad} by more than 4\%. The gap is most notable in the \textsc{List} and \textsc{Lighting} domains where \textsc{GMDP} outperforms \textsc{AdaGrad} by 14.3\% and 12.4\%, respectively, but the improvements on \textsc{Messenger} and \textsc{Workforce} are also substantial (8.4\% and 4.1\%, respectively). In the other three domains, \textsc{GMDP} and \textsc{AdaGrad} perform identically (\textsc{Container}) or demonstrate differences of up to 2\% (\textsc{Calendar} and \textsc{File}). Interestingly, for the \textsc{Calendar} domain, performing GMDP training without the features and filtering  (\textsc{GMDP$-$FA}) yields the best accuracy.


\paragraph{Features and Application Logic}

Removing our new features (\textsc{GMDP$-$F} and \textsc{AdaGrad$-$F}) yields an averaged accuracy decrease of 17.6\% (from 44.5\% to 26.9\% for \textsc{GMDP})  or 6.7\% (from 39.1\% to 32.4\% for \textsc{AdaGrad}).  Removing the application logic filtering yields an averaged decrease of 10.0\% (\textsc{GMDP} vs. \textsc{GMDP$-$A}) or 3.3\% (\textsc{AdaGrad} vs. \textsc{AdaGrad$-$A}). Finally, removing both the features and the application logic yields an additional average degradation: a total of 19\% for \textsc{GMDP} and of 10.8\% for \textsc{AdaGrad}.

\paragraph{Combined Effect}

The impact of our features and application logic filtering is much more substantial for  \textsc{GMDP} models, both on average and for the four domains where \textsc{GMDP} outperforms \textsc{AdaGrad} (four rightmost domain columns of the table). Particularly, while in these four setups the combination of GMDP training with the features and filtering yields a substantial improvement over AdaGrad training with these additions, excluding either the features or the filtering often results in an advantage for AdaGrad training. In the other three setups, where \textsc{GMDP} and \textsc{AdaGrad} perform similarly, excluding the features or the filtering yields similar effects (with the exception of \textsc{GMDP$-$FA} in the \textsc{Calendar} domain).

This observation provides an important insight about our modeling decisions. At their best, our three zero-shot components provide a complementary effect and these are the cases where our full model, GMDP, is most useful. When this complementary effect is not observed, AdaGrad and GMDP training are equally effective.

\paragraph{Comparison with In-domain Training}

In order to better quantify the impact of zero-shot training, we report results for AdaGrad training in the in-domain setup, i.e. when the parser is trained with the target domain's training set. Note that while the zero-shot models are trained with 603-611 training examples (all the training examples of the source domains), the in-domain models are trained with 96-104 examples only.

As shown in Table \ref{tab:MainResults}, the accuracy of \textsc{AdaGrad} with in-domain training is 15.2\% higher than that of \textsc{GMDP} with zero-shot training (59.7\% vs. 44.5\%), despite the smaller number of training examples. A comparison between \textsc{AdaGrad} and \textsc{AdaGrad$-$FA} reveals that in the in-domain setup, our new features and filtering logic yields only a modest performance gain that corresponds to 3.2\% on average (59.7\% vs. 56.5\%). This is another induction for the relevance of our zero-shot components to zero-shot adaptation.

\paragraph{Error analysis.}
For each domain we sample 10 training examples per interface method, and analyze the performance of the parsers that are applied to this domain. This accumulates to 200 examples, that are used for the below error analysis.

For \textsc{GMDP}, 30.4\% of the error is in examples where the instruction is incorrect, which is the case in 33 out of the 200 sampled examples (16.5\%). 
Another major source of errors is the lexical gap between the domains. Consider the utterance \textit{remove the longest container} from the \textsc{Container} domain, with the correct logical form:
\begin{align*}
& \mathrm{removeContainers}(\mathrm{argmax}(  \\ 
& \mathbf{R}[type].\mathrm{ShippingContainer},\mathbf{R}[length])) 
\end{align*}
The word \textit{longest} did not appear in any example in the source domains, and thus none of the relevant lexicalized features associated with $\mathrm{argmax}$ were useful. In future work, we hence plan to extend our parser to take word similarity into account.

Moreover, we found that 12.8\% of the error is due to incorrect parsing of utterances that reference an entity by its index. An example of such an error is the mapping of the utterance \textit{unload the container in terminal four} to the logical form  $\mathrm{unloadContainers}(\mathbf{R}[\mathrm{length}].4)$ instead of $\mathrm{unloadContainers}(\mathbf{R}[\mathrm{index}].4)$.

\paragraph{Qualitative analysis.}

We observe that \textsc{GMDP} yields smaller weights for features that can be expected to correlate with incorrect logical forms due to the zero-shot setup. For example, in the \textsc{Container} domain annotators often referred to entities by their index (e.g. \textit{Remove the container in the third terminal}), while in the \textsc{List} domain annotators mostly refer to entities (integers) by their numeric value. When \textsc{List} is the target domain, we observe that in \textsc{GMDP} the lexicalized phrase-predicate features that indicate co-occurrence between the logical form $\mathbf{R}[\mathrm{index}]$ and phrases that do not indicate an index based reference, receive smaller weights when compared to \textsc{AdaGrad}. 
For example, we find that the feature that indicates a co-occurrence between $\mathbf{R}[\mathrm{index}]$ and the phrase \textit{in} corresponds to the largest decrease in weight percentile rank: 88.7 points. At the same time, the feature that indicates a co-occurrence with the phrase \textit{the first} (which should correlate with $\mathbf{R}[\mathrm{index}]$ being in the logical form) corresponds to the largest increase in weight percentile rank: 86.1 points. 

As a result of this change in feature weighting, for \textsc{List} utterance such as: \textit{Remove the number 2 from the list}, \textsc{GMDP} tends to yield correct logical forms  (e.g. $\mathrm{remove} (\mathbf{R}[\mathrm{value}.2])$), unlike \textsc{AdaGrad} that tends to query entities by their index instead of by their value \mbox{(e.g. $\mathrm{remove} (\mathbf{R}[\mathrm{index}.2])$).} 


\section{Conclusion}

We presented a novel task of zero shot semantic parsing for instructions, and introduced a new dataset. We proposed a new training algorithm as well as features and filtering logic that should enhance zero-shot learning, and integrated them into the FParser \cite{pasupat2015compositional}. Our new parser substantially outperforms the original parser and we further show that each of our zero-shot components is vital for this improvement.

%
%

We hope this work will inspire readers to use our framework for collecting a larger dataset and experimenting with more approaches. Our framework is designed to allow the definition of new domains and collecting examples with minimal effort. Promising future directions include experimenting with our zero-shot adaptation methods in the context of neural semantic parsing (after increasing the number of examples per domain) and extending the dataset to include more complicated applications and multi-utterance instructions.

\section*{Acknowledgements}

We would like to thank the members of the IE@Technion NLP group for their valuable feedback and advice. This research has been funded by an ISF personal grant on "Domain Adaptation in NLP: Combining Deep Learning with Domain and Task Knowledge".

\bibliography{naaclhlt2019}
\bibliographystyle{acl_natbib}


\appendix

\section {Hyper-parameter Tuning}

The hyper-parameters we consider in the grid search are as follows. L1 regularization coefficient: $\{0.001, 0.01\}$; initial step-size: $\{0.01, 0.1\}$; number of training iterations (for the \textsc{GMDP} algorithm, the number of training iterations in the second step): $\{1,2,3\}$. 
For the \textsc{GMDP} algorithm, the grid search also included the number of training domains used for the first step ($D_1$): $\{3, 4\}$ (the remaining domains are used in the second step, $D_2$) with three random domain orderings for determining the $D_1,D_2$ partition (see the last paragraph of Section 4.2.1 for how these hyper-parameters are used), and the number of training iterations during the first step: $\{2, 4\}$.

\end{document}